\documentclass{article}

\usepackage{PRIMEarxiv}

\usepackage[utf8]{inputenc} % allow utf-8 input
\usepackage[T1]{fontenc}    % use 8-bit T1 fonts
\usepackage{hyperref}       % hyperlinks
\usepackage{url}            % simple URL typesetting
\usepackage{booktabs}       % professional-quality tables
\usepackage{amsmath,amssymb,amsfonts}       % blackboard math symbols
\usepackage{nicefrac}       % compact symbols for 1/2, etc.
\usepackage{microtype}      % microtypography
\usepackage{lipsum}
\usepackage{fancyhdr}       % header
\usepackage{graphicx}       % graphics

\title{Fact or Artifact? Revise Layer-wise Relevance Propagation on various ANN Architectures\thanks{This work was supported by the Helmholtz School for Marine Data Science (MarDATA) funded by the Helmholtz Association (Grant HIDSS-0005).}}

\author{
    \textbf{Marco Landt-Hayen} \\
    \textit{GEOMAR Helmholtz Centre for Ocean Research} \\
    Kiel, Germany \\
    mlandt-hayen@geomar.de \\
    \and
    \textbf{Willi Rath} \\
    \textit{GEOMAR Helmholtz Centre for Ocean Research} \\
    Kiel, Germany \\
    wrath@geomar.de \\
    \and
    \textbf{Martin Claus} \\
    \textit{Christian-Albrechts-Universität} \\
    Kiel, Germany \\
    mclaus@geomar.de \\
    \and
    \textbf{Peer Kröger} \\
    \textit{Christian-Albrechts-Universität} \\
    Kiel, Germany \\
    pkr@informatik.uni-kiel.de
    }

\begin{document}
\maketitle

\begin{abstract}
Layer-wise relevance propagation (LRP) is a widely used and powerful technique to reveal insights into various artificial neural network (ANN) architectures. LRP is often used in the context of image classification. The aim is to understand, which parts of the input sample have highest relevance and hence most influence on the model prediction. Relevance can be traced back through the network to attribute a certain score to each input pixel. Relevance scores are then combined and displayed as heat maps and give humans an intuitive visual understanding of classification models. Opening the black box to understand the classification engine in great detail is essential for domain experts to gain trust in ANN models. However, there are pitfalls in terms of model-inherent artifacts included in the obtained relevance maps, that can easily be missed. But for a valid interpretation, these artifacts must not be ignored. Here, we apply and revise LRP on various ANN architectures trained as classifiers on geospatial and synthetic data. Depending on the network architecture, we show techniques to control model focus and give guidance to improve the quality of obtained relevance maps to separate facts from artifacts.
\end{abstract}


\section{Introduction \label{sec:Introduction}}
Image classification refers to assigning one or more class labels to a two-dimen-sional image. Here, we focus on binary classification problems and have only two distinct classes. Instead of having a discrete class label, one often assigns some continuous target value to each sample. The class label is then derived from the target value by defining certain thresholds. Like this, the initial classification problem becomes a regression problem. The task is then to predict some continuous target value from all input pixels. Generally, the relationship between inputs and targets is highly nonlinear. Artificial neural networks (ANNs) are state-of-the-art for this task. In this work, we will revise the use of various ANN architectures for image classification. In particular, we work with multilayer perceptrons (MLPs), convolutional neural networks (CNNs) and Echo State Networks (ESNs). MLP models have frequently been used for image classification \cite{Toms2020,Coskun2003,Shubathra2020}. CNN architectures aim to detect objects and structures in the underlying samples and are powerful tools for image classification \cite{Krizhevsky2017,Kadam2020}, object detection \cite{Mottaghi2014} or semantic segmentation \cite{Redmon2016}. With their inherent shared weights philosophy, CNNs need less trainable parameters, compared to MLPs. ESNs differ from MLPs and CNNs, as they are a special type of recurrent neural networks (RNNs), originally designed for time series forecasting \cite{Kim2020,Gallicchio2017}. To use two-dimensional images as inputs, samples may be sliced column- or row-wise, respectively. Like this, either the $x-$ or the $y-$dimension is turned into a time dimension, as done in \cite{LandtHayen2022}.

As input data, we use real world geospatial data. In particular, we choose a well-understood Earth System Variability, the El Ni\~{n}o Southern Oscillation (ENSO). The current ENSO phase can be detected from spatially averaged sea surface temperature (SST) anomalies in the so-called Ni\~{n}o 3.4 region in the Tropical Pacific. But ENSO is a complex phenomenon and is related to climate anomalies over large distances, also referred to as teleconnections. ENSO leaves its footprint outside the Tropical Pacific, e.g. at the African West coast near Angola \cite{Shannon1987}. Here, we use ANN models as ENSO detectors on two-dimensional samples of SST anomalies. The two classes are either "El Ni\~{n}o" or "La Ni\~{n}a", which can be associated with unusual warm or cold sea surface temperature in the Ni\~{n}o region, respectively.

While all models perfectly perform on this simple classification task, our aim is \textit{not} to find a superior classifier. Here, we are interested in getting a deeper understanding of how the models come to their conclusions and explain the reasoning of the models' classification engine by using ENSO as a toy problem. A variety of techniques from the domain of explainable artificial intelligence (xAI) exists to open the black box of ANNs. In backward optimization \cite{Olah2017}, the inputs for a trained model are modified to maximize the network's confidence in the obtained output. The goal is to generate some optimal input. A saliency map \cite{Simonyan2013} aims to highlight areas in a given input sample that show strongest support towards a given class. Other techniques compute the gradient of a prediction with respect to the input pixels in a sensitivity analysis \cite{Nourani2012}, to reveal insights in how sensitive the model output depends on slight modifications of input values. Our purpose is more general. We want to find each input pixel's contribution to the model output. This can be achieved by layer-wise relevance propagation (LRP). The concept of LRP has been introduced by Bach et al. \cite{bib:Bach2015}. The model prediction is taken as final relevance and is then traced back through all layers until reaching the input space to assign individual relevance scores to each input pixel. LRP has been extended from a practical point of view by Montavon  et al. \cite{bib:Montavon2017}. In their work, they provide a set of concrete propagation rules for tracing relevance back through the layers.

Toms et al. \cite{Toms2020} recently applied LRP to MLP models trained as ENSO detector. They show mean relevance maps obtained from all El Ni\~{n}o samples. Significant relevance is only found in the Ni\~{n}o region, as one might expect on the first sight. There are no other spots of high relevance outside this area. In our earlier work \cite{LandtHayen2022}, we proposed to use ESN models on the same task and also performed LRP on ENSO. We found mean relevance maps for El Ni\~{n}o samples with a far more subtle structure, also highlighting Ni\~{n}o region but highest relevance is found on the passage between South Africa and Antarctica.

Moreover, the leak rate in the reservoir transition of an ESN determines the reservoir memory and hence is a crucial parameter \cite{LandtHayen2022}. Here, we give a heuristic for setting the leak rate in relation to the number of time steps. 

In addition to MLP and ESN models, we also apply LRP to CNN models in this work. For MLPs and CNNs used for image classification, regularization of weights is found to be essential. We can force MLP and CNN models to focus only on a very narrow spot in the Tropical Pacific to discriminate El Ni\~{n}o from La Ni\~{n}a, by driving small weights to zero to favor sparse weight matrices in the training process \cite{Krogh1991}.

To get a deeper understanding, we step back and first apply MLP, CNN and ESN models to synthetic samples, where we exactly know the relationship between inputs and outputs by design. Only with a sound understanding of pitfalls related to LRP, we can use this powerful technique on unknown problems to help domain experts to explain their models, gain trust in models' predictions and avoid misinterpretations. Like this, LRP bears good prospects to help finding new relationships and understanding unknown teleconnections \cite{Pak2014,Park2018,Zhang2019}.

Our main contributions are as follows:
\begin{itemize}
\setlength\itemsep{.2em}
\item We revise LRP on various ANN architectures used for image classification.
\item We show how to control the model focus of MLP and CNN networks by weight regularization.
\item For ESNs, we find the leak rate to be the crucial parameter and give a heuristic to choose it appropriately.
\item Some of the additional spots of high relevance found in mean relevance maps from ESN models are identified as artifacts. Furthermore, we propose a technique to erase these artifacts.
\end{itemize}

The rest of this work is structured as follows: In Section 2 we briefly introduce synthetic and real world data used for our experiments. Section 3 outlines the concept of LRP customized to MLP, CNN and ESN models. The application of LRP to synthetic and real world ENSO patterns is presented in Section 4. A detailed discussion and conclusion is found in Section 5, followed by technical details, remarks on availability of data and annotated code in the Appendix.

\section{Data \label{sec:Data}}
\begin{figure}[!b]
\includegraphics[width=\textwidth]{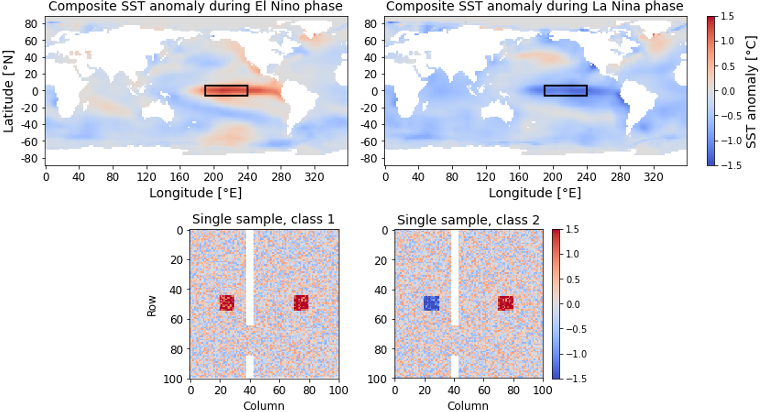}
\caption{Upper part: Composite average SST anomaly patterns for El Ni\~{n}o (left-hand side) and La Ni\~{n}a (right-hand side) events. A black rectangle highlights Ni\~{n}o 3.4 region. Lower part: Exemplary samples for classes 1 (left-hand side) and 2 (right-hand side) of synthetic data.}
\label{fig:Data}
\end{figure}

In this section, we introduce SST anomaly fields related to ENSO used as real world data. After that, we describe the design of our synthetic samples. In this work, we use two-dimensional monthly mean SST anomaly fields for the years 1880 through 2021 as real world inputs. Raw SST data is provided by the NOAA Physical Sciences Laboratory \cite{NOAAdownload}. Input samples consist of $89 \times 180$ grid points on a 2° by 2° latitude-longitude grid. Several indices are used to monitor ENSO activity based on SST anomalies in the Tropical Pacific \cite{NCARclimateguide}. Morrow et al. \cite{Morrow2010} define in their work an index from the Ni\~{n}o 3.4 region (5°N–5°S, 120–170°W), which we use as target in this work. El Ni\~{n}o and La Ni\~{n}a events are characterized by positive and negative SST anomalies in the Ni\~{n}o 3.4 region, referred to index values $\ge 0.5$ and $\le -0.5$, respectively. Neutral states in between, are not considered for classification, here. The total number of 1,041 samples is split, using the first 80\% as training data (832 samples) and the remaining 20\% for validation (209 samples). The upper part of Figure \ref{fig:Data} shows composite average SST anomaly fields for El Ni\~{n}o and La Ni\~{n}a.

As synthetic data, we create quadratic samples of dimension $100 \times 100$ pixels and define two distinct classes, class 1 and class 2, respectively. On the right-hand side of each sample, we have a vertically centered square of value +1, identical for both classes. The left-hand side also contains a vertically centered square of values +1 and -1 for classes 1 and 2, respectively, as equivalent for SST anomalies in the Ni\~{n}o region in real world samples. Classes can be uniquely discriminated by the square on the left-hand side, while the other only serves as an additional but irrelevant feature. We add random noise to all samples, drawn from a uniform distribution in the interval $[-0.5, 0.5]$. In addition to that, we include some vertical regions of invalid grid points with a small gateway to all samples, similar to land masses in real world samples, where SST anomalies are not defined. Target values are +1 and -1 for classes 1 and 2, respectively. Training data consists of 400 samples, 200 for each class, while validation data consists of 100 samples, 50 for each class. The lower part of Figure \ref{fig:Data} shows single samples for both classes as example.

\section{Methods and Models\label{sec:Models}}

In this section, we briefly recap the ANN architectures and sketch the concept of LRP customized to MLP, CNN and ESN models. MLPs are feedforward networks and consist of a specified number of layers. The input layer is connected to the input samples. This requires the number of input units to equal the number of input pixels. Two-dimensional samples are therefore transformed into a one-dimensional vector. CNNs are also feedforward networks. The convolution is done by specified kernels scanning the input image and producing so-called feature maps. After a desired number of convolutions and optionally pooling operations, we obtain the final feature maps. Here, we flatten these final feature maps and stack fully connected dense layers on top of the underlying CNN part. The basic form of an ESN model consists of an input and an output layer and a reservoir of sparsely connected units in between. Once randomly initialized, the input and reservoir weights and biases are kept fixed. We then feed a certain number of input features for a specified number of time steps into the model and record the final reservoir states. Only the output weights and bias are trained by regressing final reservoir states onto targets. There is no backpropagation of errors in the training process, as for MLPs or CNNs. Equation \ref{equation:first_reservoir_state} states the initial reservoir states $x(t=1)$ for our ESN model. We have a leaky reservoir with leak rate $\alpha \in [0,1]$, as discussed in \cite{Jaeger2007}. For larger leak rates, the reservoir states react faster to new inputs. $u(t)$ denote current time step's inputs, $W_{in}$ and $b_{in}$ are input weights and biases, respectively.

\begin{equation}
x(t=1) = \alpha \, act[W_{in} u(t=1) + b_{in}]
\label{equation:first_reservoir_state}
\end{equation}

Reservoir state transition for subsequent time steps is outlined in Equation \ref{equation:reservoir_state_transition}. A fraction $(1-\alpha)$ of the previous reservoir states $x(t-1)$ is kept and combined with the term inside the activation $act(.)$, here \emph{tanh}. $W_{res} x(t-1) + b_{res}$ denotes the recurrence inside the reservoir, with reservoir weights and biases $W_{res}$ and $b_{res}$, respectively.

\begin{equation}
    x(t) = (1-\alpha)\,x(t-1) + \alpha \, act[W_{in} u(t) + b_{in} + W_{res} x(t-1) + b_{res}]
  \label{equation:reservoir_state_transition}
\end{equation}

Toms et al. \cite{Toms2020} describe in their work, how LRP can be applied to MLP and CNN models, as both are feedforward networks. The model output $Y$ is taken as final relevance and is then traced back through all layers until reaching the input space to assign relevance scores $R_n^{(1)}$ to each of the $n$ input pixels, as stated in Equation \ref{equation:LRP_1}.
\begin{equation}
Y = \sum_{n} R_n^{(1)}
\label{equation:LRP_1}
\end{equation}

Starting with model output $Y$, relevance is distributed back from the output layer through lower layers until we reach the input layer. To preserve total relevance in each layer, we have a second constraint, as stated in Equation \ref{equation:LRP_2}.

\begin{equation}
Y = ... = \sum_j R_j^{(l+1)} = \sum_i R_i^{(l)} = ... = \sum_{n} R_n^{(1)}
\label{equation:LRP_2}
\end{equation}

Samples of class 1 have positive target values by design. For this reason, we consider only positive contributions of pre-activations as propagation rule. Assume we have $j$ units in layer $(l+1)$ with known relevance scores $R_{j}^{(l+1)}$, the relevance $R_{i=i_0}^{(l)}$ for unit $i_0$ of layer $(l)$ is obtained by Equation \ref{equation:LRP_propagation}, using $z_{ij}^+ = max(a_iw_{i:j},0)$. Here, $a_i$ denotes activation of units $i$ of layer $(l)$ and $w_{i:j}$ denotes the weight connecting unit $i$ from layer $(l)$ with unit $j$ from layer $(l+1)$. For class 2 samples, we have negative targets. We start with absolute prediction values as final relevance, since relevance is defined to be positive, and only consider negative contributions of pre-activations.

\begin{equation}
R_{i=i_0}^{(l)} = \sum_j \frac{z_{i_0j}^+}{\sum_i z_{ij}^+} R_{j}^{(l+1)}
\label{equation:LRP_propagation}
\end{equation}

This LRP concept needs to be modified for our ESN models, as described in great deatail in \cite{LandtHayen2022}, since we have a recurrence in time. Reservoir transition needs to be unfolded. Each time step is equivalent to one layer. A fraction of the total relevance is attributed to each time step's inputs. Remaining relevance is passed on until we reach the initial input. Finally, all relevance scores attributed to all time step's inputs can be composed to obtain a relevance map with same dimensions as the input sample.

\section{Results \label{sec:Results}}

\begin{figure}[h]
\includegraphics[width=\textwidth]{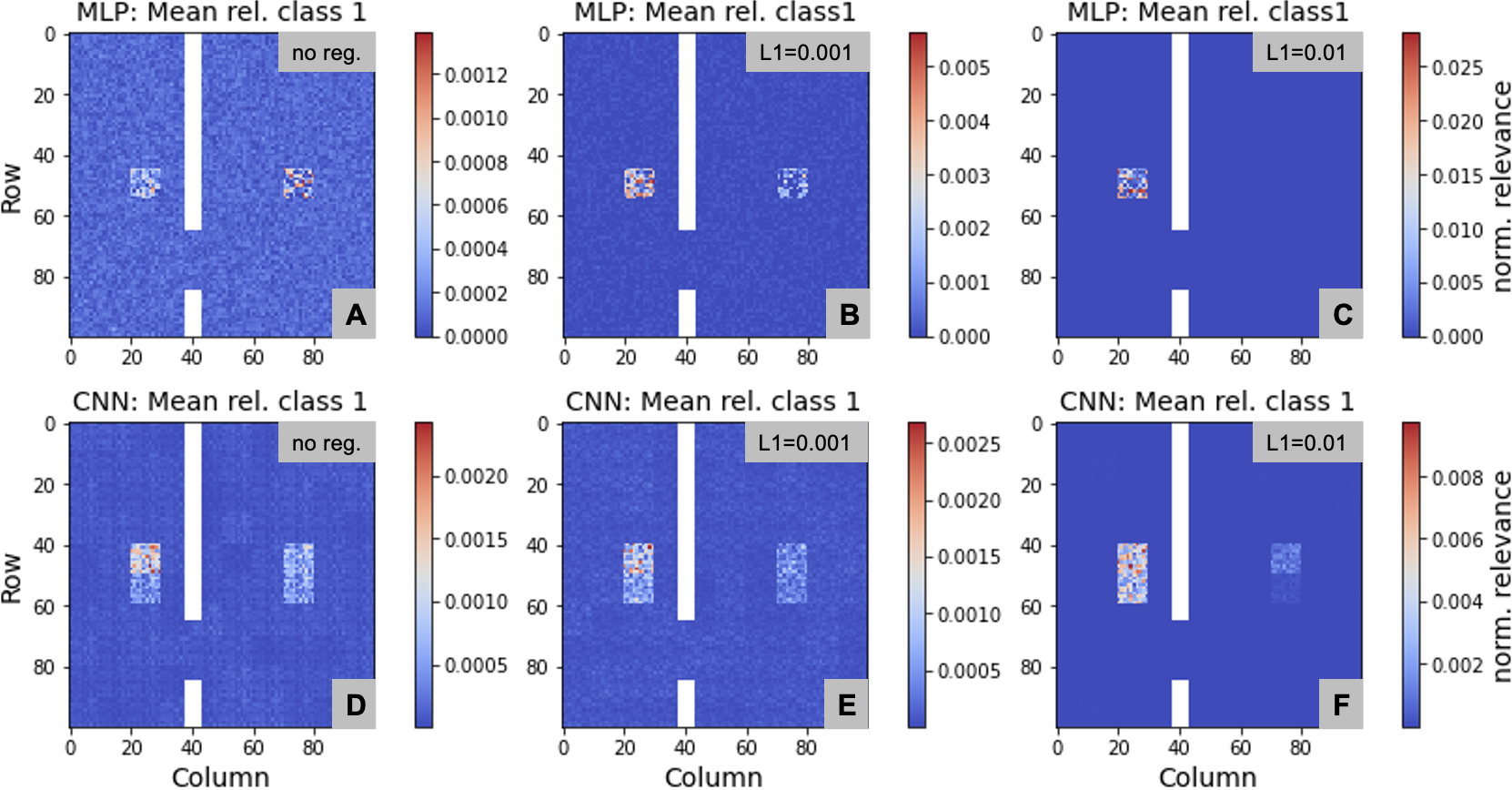}
\caption{Mean relevance maps for MLP (upper part) and CNN (lower part) on class 1 for different weight regularization. A, B, C: MLP without regularization, L1=0.001 and L1=0.01, respectively. D, E, F: CNN without regularization, L1=0.001 and L1=0.01, respectively.}
\label{fig:LRP_MLP_CNN_class1}
\end{figure}

In this section, we present results from LRP experiments with MLP, CNN and ESN models on synthetic and real world data. By design, we know that our synthetic samples can uniquely be discriminated by looking at the left square. Figure \ref{fig:LRP_MLP_CNN_class1} shows mean relevance maps for class 1 obtained from MLP (upper part) and CNN models (lower part) with and without regularization of hidden layer's weights. Without regularization, we find blurry mean relevance maps for both models with highest relevance on both squares. By adding regularization terms, we successfully force both models to focus on the left square. For the CNN we find artifacts in form of quadratic patches in mean relevance maps.

\begin{figure}[!b]
\includegraphics[width=\textwidth]{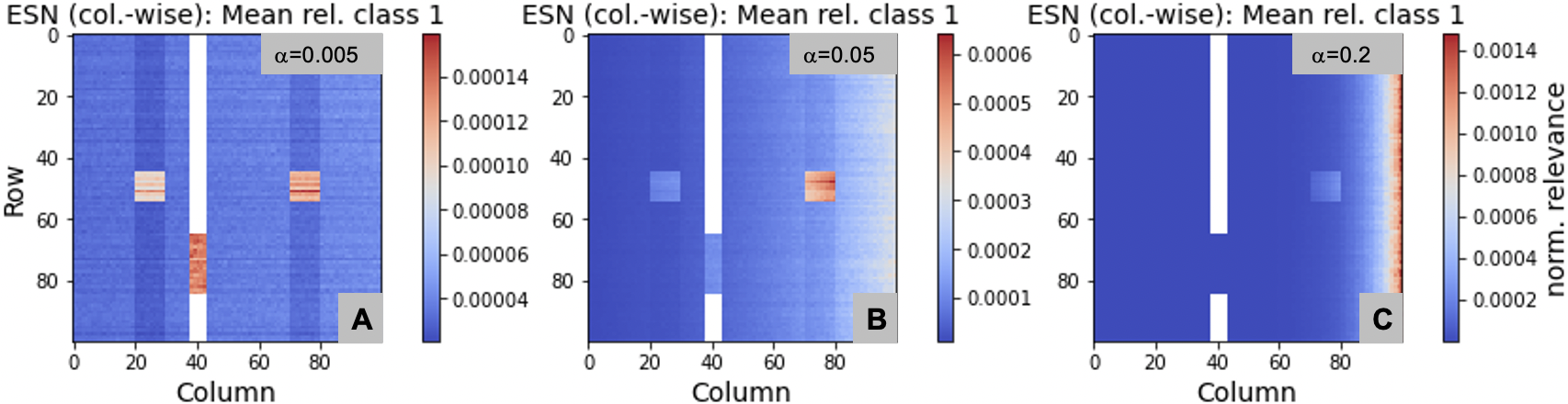}
\caption{Mean relevance maps for ESN on class 1, feeding samples column-wise into the model. Fading memory experiment with various leak rates. A: $\alpha=0.005$, B: $\alpha=0.05$, C: $\alpha=0.2$.}
\label{fig:LRP_ESN_colsiwe_class1}
\end{figure}

For ESN models, only the output weights are trained by regressing final reservoir states onto targets. This limits the use of weight regularization. However, for ESN models the leak rate $\alpha$ determines reservoir dynamics and memory. Figure \ref{fig:LRP_ESN_colsiwe_class1} shows mean relevance maps for ESN models, feeding class 1 samples column-wise into the model, starting with the leftmost column. For $\alpha=0.005$, we find an equal amount of summed relevance over both squares. But highest relevance is found on the gateway in the barrier of invalid grid points. With increasing leak rate ($\alpha=0.05$) the reservoir's memory starts to fade, summed relevance on the left square is decreased. Still, the model succeeds to classify all samples correctly. At some point ($\alpha=0.2$) the reservoir loses its memory of the left half. All relevance is bundled rightmost and classification accuracy for validation samples drops to random guessing. In this case, $\alpha=0.005$ is a suitable choice for the leak rate, since it guarantees all time steps' inputs to be equally considered.

\begin{figure}[h]
\includegraphics[width=\textwidth]{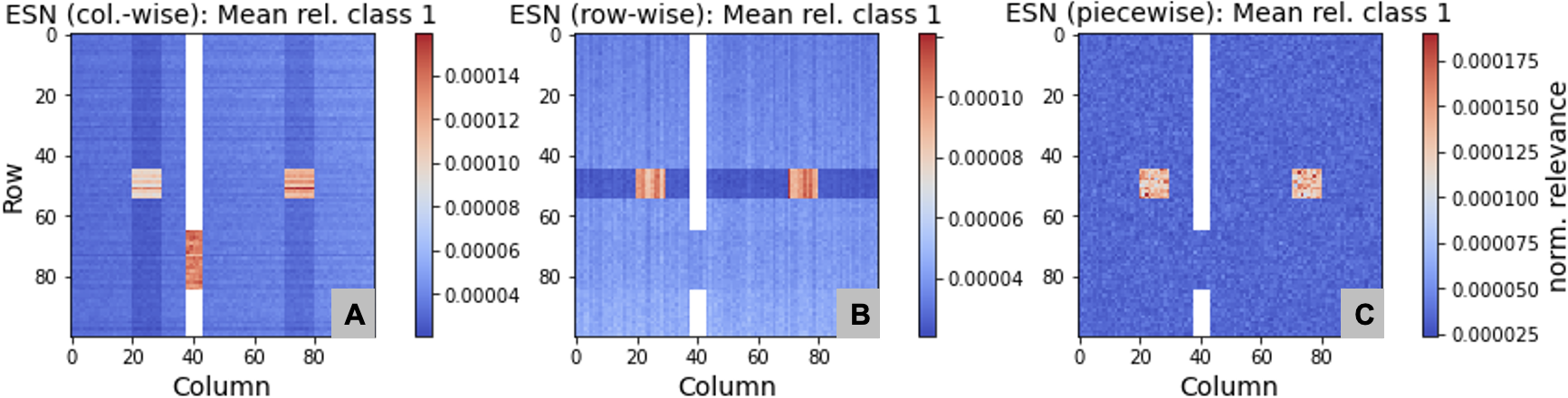}
\caption{Mean relevance maps for class 1 samples from ESN models using different techniques to feed two-dimensional inputs into the models. A: column-wise, B: row-wise, C: piecewise.}
\label{fig:LRP_ESN_all_class1}
\end{figure}

\begin{figure}[!b]
\includegraphics[width=\textwidth]{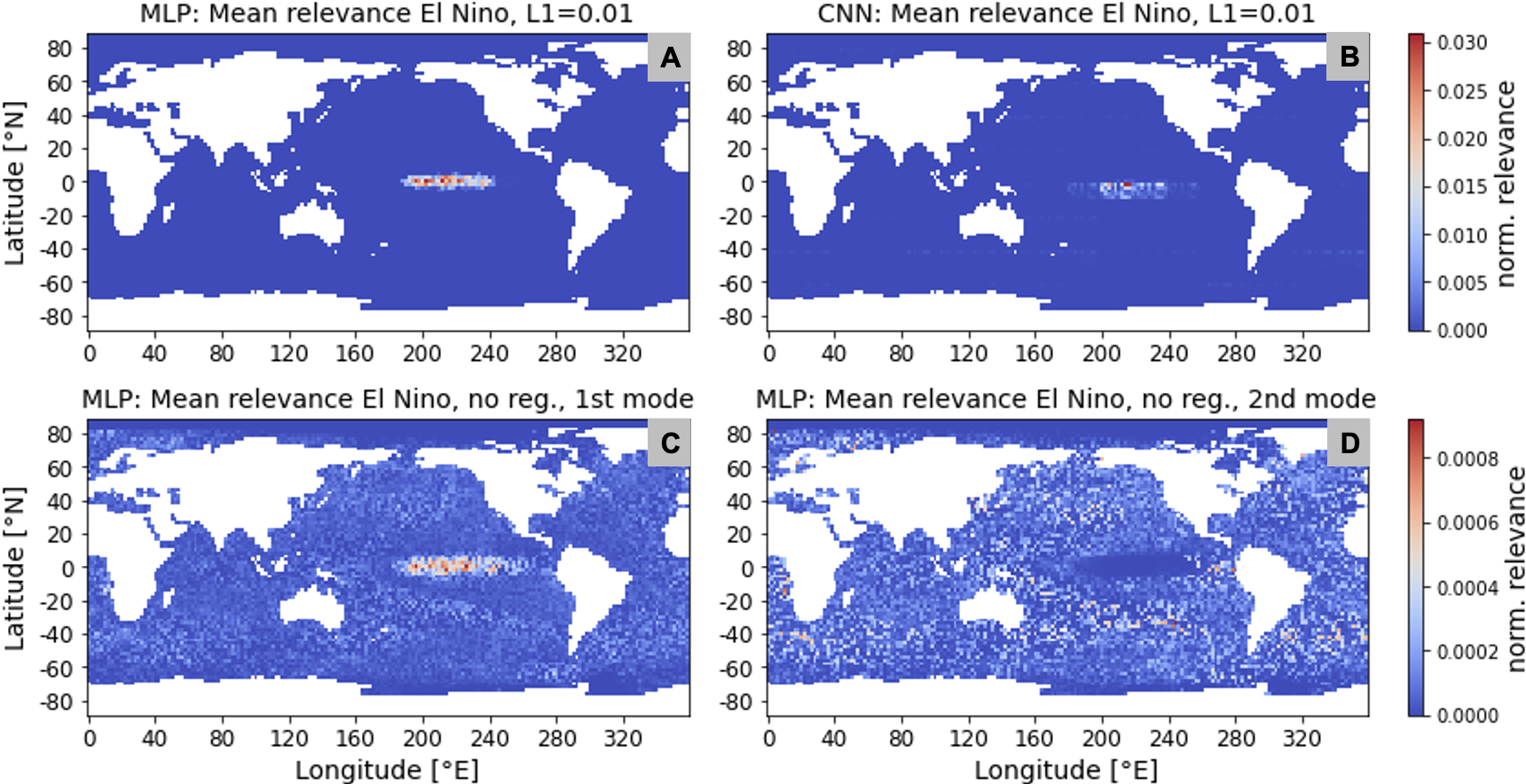}
\caption{Mean relevance maps for El Ni\~{n}o using MLP and CNN models with different weight regularization. A: MLP, L1=0.01, B: CNN, L1=0.01, C and D: MLP 1st and 2nd mode, respectively, no regularization.}
\label{fig:LRP_MLP_CNN_real_world}
\end{figure}

With that choice of the leak rate, we compare various techniques to use two-dimensional input samples for ESN models. Figure \ref{fig:LRP_ESN_all_class1} shows mean relevance maps for ESN models, feeding class 1 samples column- or row-wise into the model. We find horizontal and vertical stripes, respectively. We don't find increased relevance on the gateway when feeding samples row-wise into the model. As third approach, we propose to split input samples into equal-sized pieces, considering only valid grid points. The resulting mean relevance map for class 1 samples is also shown in Figure \ref{fig:LRP_ESN_all_class1}. We find a perfectly equal amount of summed relevance over both squares. Besides that, there are no artifacts in form of stripes and we don't find any increased relevance on the gateway.

Switching to real world samples, the upper part of Figure \ref{fig:LRP_MLP_CNN_real_world} shows mean relevance maps for El Ni\~{n}o, obtained from the MLP and CNN model with regularization (L1=0.01). We find strong and only focus inside Ni\~{n}o region. The lower part of Figure \ref{fig:LRP_MLP_CNN_real_world} shows MLP results without regularization. We find two complementary modes on the exact same parameter setting.

\begin{figure}[!b]
\includegraphics[width=\textwidth]{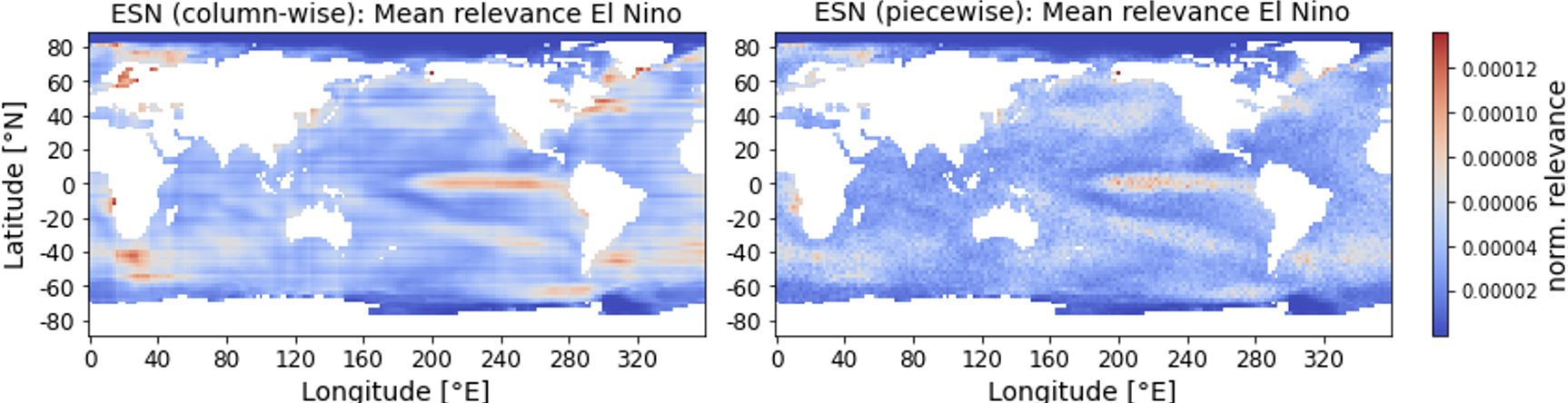}
\caption{Mean relevance maps for El Ni\~{n}o from ESN models using different techniques to feed inputs into the model. A: column-wise, B: piecewise.}
\label{fig:LRP_ESN_real_world}
\end{figure}

Figure \ref{fig:LRP_ESN_real_world} compares mean relevance maps for El Ni\~{n}o, obtained from ESN models, feeding samples column-wise and as equal-sized pieces, respectively. In both cases, we find high relevance inside the Tropical Pacific. However, feeding samples column-wise, highest relevance is found on the passage between South Africa and Antarctica and we find horizontal stripes.

\section{Discussion and Conclusion \label{sec:Conclusion}}

In this section, we will discuss all results from the previous section in detail and give an outlook on future research. We have shown, that our MLP and CNN models can be successfully forced to focus mainly on the left square of our synthetic samples, which is sufficient to fulfill the classification task. For our CNN model we find artifacts in form of quadratic patches in mean relevance maps, due to shared kernel weights. When it's L1 weight regularization found to be the key to force MLP and CNN models to focus, we find the leak rate to be crucial for ESN models to determine reservoir memory. Leak rate needs to be chosen appropriately, to equally take all time steps' inputs into account, since we don't know in advance, which time steps contain most relevant information. Using two-dimensional inputs column- or row-wise, we find horizontal and vertical stripes, respectively, which slightly differ for individual training runs. Input samples don't show any stripes. This, and the fact, that we find stripes to depend on how we feed sample into the model, clearly unmasks these stripes to be artifacts, stemming from random initialized input weights. Highest relevance for ESN models is found on the small gateway only, when we feed samples column by column. This effect of \emph{squeezed} relevance is caused by the varying number of valid grid points for different columns. The number of valid grid points per row also varies, but the effect is much smaller and can hardly be recognized. As a solution, we propose a new technique of considering only valid grid points. These grid points are transformed to a one-dimensional vector, permuted and split into equal-sized pieces. Like this, stripes disappear and we no longer find elevated relevance on the gateway. However, both squares are found to be equally relevant, since we cannot apply weight regularization on ESN models.

From a practical point of view, the model choice in combination with parameter setting is a tradeoff between explainability and performance. Table \ref{tab:evaluation} shows evaluation metrics for selected models trained on synthetic samples. While all models perfectly perform the classification task with $100\%$ accuracy, the mean squared error (mse) of model predictions compared to true targets and the model focus varies. MLP with weight regularization shows lowest mse and focuses completely on the left square. This leads to high performance in terms of efficiently discriminating classes but low explainability, since all features with minor relevance are suppressed. The ESN model with feeding samples as equal-sized pieces comes with a higher mse on validation samples compared to MLP and CNN models and thus lower performance on the classification task. However, mean relevance maps from ESN models highlight all features involved in the underlying input samples, that could reveal valuable insights into existing teleconnections and thus offer higher explainability.

\begin{table}
\caption{Evaluation metrics for selected models trained on synthetic samples: mse on training and validation data, the ratio of summed mean relevance over both squares in relation to total relevance, the relative amount of summed mean relevance over the left square in relation to both squares, and the number of trainable parameters.}\label{tab:evaluation}

\begin{tabular}{|c|c|c|c|c|}
\hline
& \textbf{MLP} no reg. & \textbf{MLP} L1=0.01 & \textbf{CNN} L1=0.01 & \textbf{ESN} piecewise \\
\hline
$mse_{train}$ & 0.01 & 0.001 & 0.005 & 0.001 \\
\hline
$mse_{val}$ & 0.053 & 0.002 & 0.008 & 0.180\\
\hline
$\sum\limits_{both\,sq}rel \,\,/ \sum\limits_{total}rel$ & 7\% & 92\% & 40\% & 7\%   \\
\hline
$\sum\limits_{left\,sq}rel \,\,/ \sum\limits_{both\,sq}rel$ & 48\% & 100\% & 98\% & 50\% \\
\hline
trainable parameters & 19,205 & 19,205 & 2,223 & 301 \\
\hline
\end{tabular}
\end{table}

Trained on real world data, our MLP and CNN models with weight regularization act as a very efficient classifier and only focus on a narrow spot in the Ni\~{n}o region. Without regularization, the MLP model reveals more structure in mean relevance maps. Moreover, we find two complementary modes, due to random initialization of weights and biases and the stochastic nature of the learning algorithm. The first mode shows strong focus on the Ni\~{n}o region, while the second mode completely ignores the very same region. With this second mode, the MLP still successfully discriminates ENSO patterns by encountering information from outside Tropical Pacific.

Using real world samples column-wise for training our ESN model, we again find horizontal stripes in the obtained mean relevance map, that have already been identified as artifacts in experiments on synthetic data. Highest relevance is found in the region between South Africa and Antarctica. Since SST input data is only defined over the ocean, the number of valid grid points is smallest for these columns, forcing relevance to be compressed onto the remaining valid grid points. Again, splitting input samples into equal-sized pieces erases both artifacts in mean relevance maps obtained from ESN models. Like this, our proposed method enhances the explainability for ESN models.

Overall, neither model is found to be superior for image classification with LRP. While MLP and CNN models appear to be \textit{efficient} in focusing on most relevant features, mean relevance maps of ESNs reveal valuable information on existing teleconnections. It is also worth to mention that ESN models come with a significantly lower number of trainable parameters, allowing these models to be trained on less training data without overfitting. With this advantage, ESNs have good prospects for further use on geospatial data and could also be used in combination with LRP in the context of time series prediction, when we replace two-dimensional inputs by a number of e.g. climate indices. This approach could then serve as alternative for long short-term memory (LSTM) models with attention mechanisms. 

\section*{Appendix\label{sec:Appendix}}

In this work we only show reproducible results. All models have been implemented in Python (version 3.9.9) using Tensorflow (version 2.4.1). To keep it as transparent as possible, we provide data and annotated Python code in Jupyter notebooks containing all experiments and details on models and methods \cite{GitRepo}.

\section*{Supplementary Material\label{sec:SuppMat}}

\subsection*{LRP for ESNs}

In Section \ref{sec:Results} we give a heuristic on how to choose the leak rate $\alpha$ from a practical point of view. Feeding a sample with $T$ time steps into an ESN model, the aim is to equally take all time steps' inputs into account, if possible. Here, we add a mathematical validation to support the empirical approach and explain side effects related to the choice of $\alpha$. As described in Section \ref{sec:Models}, the general concept of LRP needs to be customized for ESN models. In particular, we work with leaky reservoirs. Reservoir state transition for the first time step $t=1$ and subsequent time steps $t=2..T$ is outlined in Equations \ref{equation:first_reservoir_state} and \ref{equation:reservoir_state_transition}, respectively. The leak rate $\alpha$ determines the reservoir dynamics. To compute reservoir states $x(t)$ for a given time step in the forward pass, a fraction $(1-\alpha)$ of the previous reservoir states $x(t-1)$ is kept. The current time step's inputs $u(t)$ are included in some activation term $act[W_{in} u(t) + b_{in} + W_{res} x(t-1) + b_{res}]$. This term is multiplied with $\alpha$. The larger the leak rate, the faster the reservoir reacts to new inputs.

For obtaining mean relevance maps, we unfold reservoir dynamics in time. Each time step is treated as an individual layer. We use the backward pass and start with model prediction $Y$ as final or total relevance. A portion of the total relevance is attributed to every time step's inputs. Only the remaining relevance is then traced back through the layers or time steps. Multiplication with leak rate $\alpha$ in reservoir state transition leads to an exponential decay, when tracing relevance backwards. $R(t)$ denotes residual relevance for time step $t$. Once we reach the initial time step $t=1$, all residual relevance is attributed to the initial inputs $u(t=1)$. 
\begin{equation}
R(t) = exp(-\alpha\,(T-t))
\label{equation:Residual_relevance}
\end{equation}

\begin{figure}[!b]
\includegraphics[width=\textwidth]{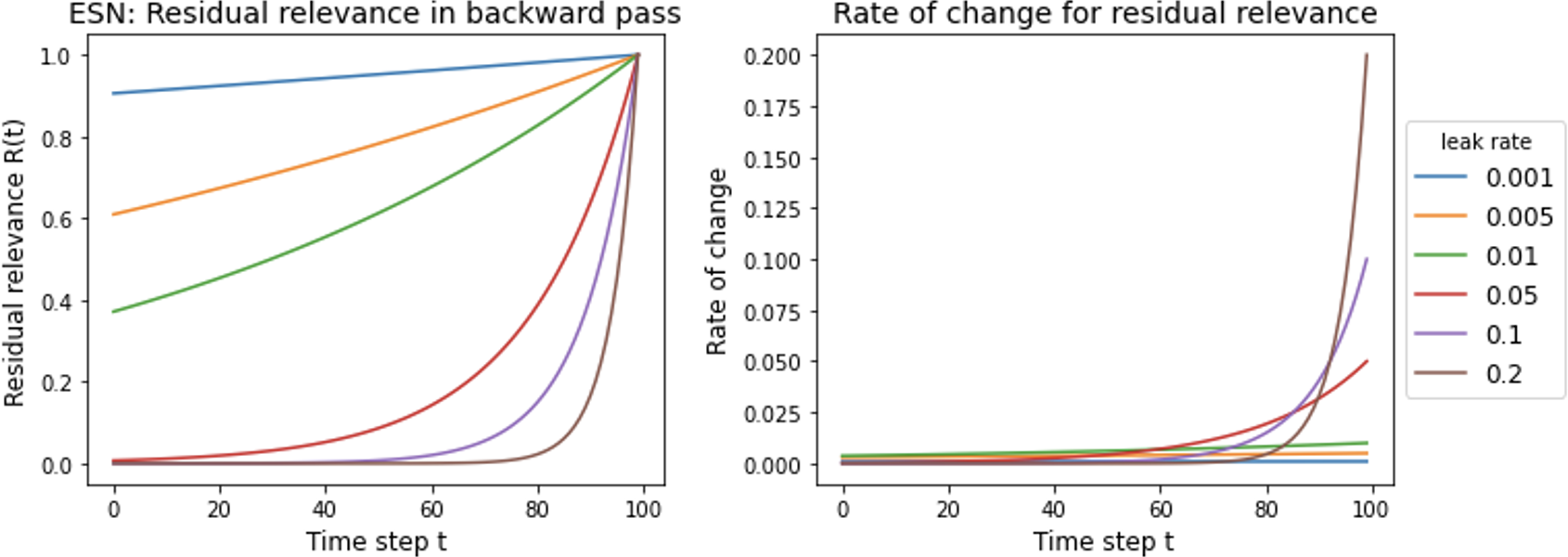}
\caption{Residual relevance approximated by an exponential decay over $T=100$ time steps for various leak rates $\alpha$ (left-hand side) and rate of change for residual relevance (right-hand side).}
\label{fig:residual_relevance_rate_of_change_ESN}
\end{figure}

Equation \ref{equation:Residual_relevance} states the the exponential decay for a given leak rate $\alpha$. The left part of Figure \ref{fig:residual_relevance_rate_of_change_ESN} shows the result for different leak rates. We clearly see the fading memory effect that has been described in Section \ref{sec:Results}. For leak rates $\alpha=0.1$ and $0.2$, only the inputs for time steps $t\geq 50$ are considered.

\begin{table}
\caption{Residual relevance $R(t=1)$ from approximation compared to empirical residual relevance $\hat{R}(t=1)$ for first time step for ESN model trained on synthetic samples with different leak rates $\alpha$.} \label{tab:residual_relevance}

\begin{center}
\begin{tabular}{|c|c|c|}
\hline
Leak rate $\alpha$ & $R(t=1)$ & $\hat{R}(t=1)$ \\
\hline
0.001 & 90.5\% & 90.7\% \\
0.005 & 60.7\% & 63.1\% \\
0.01 & 36.8\% & 41.9\% \\
0.05 & 0.7\% & 2.9\% \\
0.1 & 0.0\% & 0.1\% \\
0.2 & 0.0\% & 0.0\% \\
\hline

\end{tabular}
\end{center}
\end{table}

Table \ref{tab:residual_relevance} shows residual relevance $R(t=1)$ approximated as exponential decay according to Equation \ref{equation:Residual_relevance} for various leak rates. Here, we assume to have $T=100$ time steps. In addition to that, Table \ref{tab:residual_relevance} shows empirical residual relevance $\hat{R}(t=1)$ obtained from an ESN model trained on synthetic samples feeding samples column-wise into the model. Synthetic samples also consist of $T=100$ time steps. For leak rates $\alpha=0.001$, $0.005$ and $0.01$, we find $\hat{R}(t=1)=91\%$, $63\%$ and $42\%$ of total relevance, respectively, as residual relevance. This amount is attributed to initial inputs $u(t=1)$, which gives the initial inputs an unreasonable high relevance compared to all other time steps' inputs. For leak rates $\alpha=0.05$, $0.1$ and $0.2$, we find the remaining relevance attributed to the initial inputs to be close to zero. The exponential decay is found to be a good approximation.

Using ESN models, the aim is to equally consider all time steps' inputs, since we don't know in advance, which time steps contain most relevant information. Therefore, we need to look at the rate of change for residual relevance traced backwards through all time steps. The rate of change can be approximated as first derivative $R'(t)$ of residual relevance $R(t)$ and is stated in Equation \ref{equation:rate_of_change}.

\begin{equation}
R'(t) = \frac{\partial R}{\partial t} = \alpha\,exp(-\alpha\,(T-t))
\label{equation:rate_of_change}
\end{equation}

The right-hand side of Figure \ref{fig:residual_relevance_rate_of_change_ESN} shows the resulting rate of change for different leak rates. We find the rate of change to be almost constant for small leak rates $\alpha=0.001$, $0.005$ and $0.01$, reflecting a similar amount of total relevance attributed to each time step's inputs $u(t)$, as desired. This is violated for higher leak rates $\alpha=0.05$, $0.1$ and $0.2$, respectively. For these leak rates the ESN model puts significantly higher weights on later input steps. 

Taking residual relevance and rate of change into account, we need to find a compromise. We want all time steps to be about equally considered. This encourages us to choose small leak rates. However, we need to avoid an unreasonable high amount of residual relevance to be put on the initial time step's inputs $u(t=1)$. One possible solution is to add a dummy column of constant value one as first input time step for all input samples. Residual relevance is then absorbed by this dummy column, which doesn't affect classification in general and doesn't distort mean relevance maps, since it is identical for all samples of both classes. For showing the obtained mean relevance maps, the dummy column is then omitted.

\subsection*{Data Preprocessing}

\begin{figure}[!b]
\begin{centering}
\includegraphics[width=0.5\textwidth]{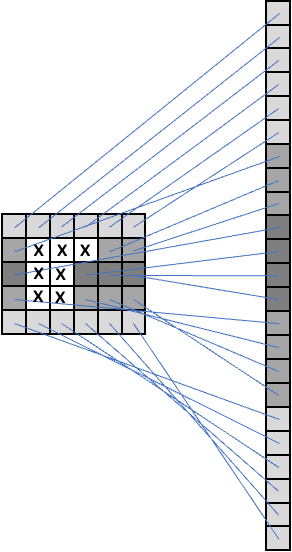}
\caption{Transformation of an exemplary two-dimensional input sample consisting of 5 rows and 6 columns into a one-dimensional vector for our MLP models. Consider only valid grid points. Invalid grid points are marked as X.}
\label{fig:preprocessing_MLP}
\end{centering}
\end{figure}

While the main body of this work focuses on models, methods and results, we provide details on data preprocessing in this section. Different ANN models require input samples to be customized in dimensionality and we need to deal with missing data at grid points located on land. For our MLP models, we only consider valid grid points of raw two-dimensional input samples. These grid points are then transformed into a one-dimensional vector as sketched in Figure \ref{fig:preprocessing_MLP}. The number of input units then equals the number of valid grid points.

\begin{figure}[h]
\begin{centering}
\includegraphics[width=0.7\textwidth]{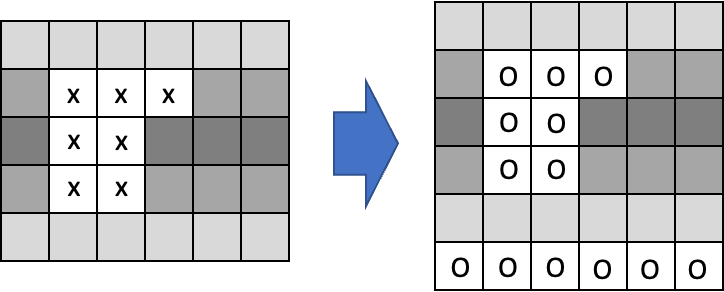}
\caption{Modification of an exemplary two-dimensional input sample consisting of 5 rows and 6 columns by replacing missing values and adding an additional row of zeros for our CNN models. Invalid grid points and zeros are marked as X and O, respectively.}
\label{fig:preprocessing_CNN}
\end{centering}
\end{figure}

\begin{figure}[!b]
\begin{centering}
\includegraphics[width=0.9\textwidth]{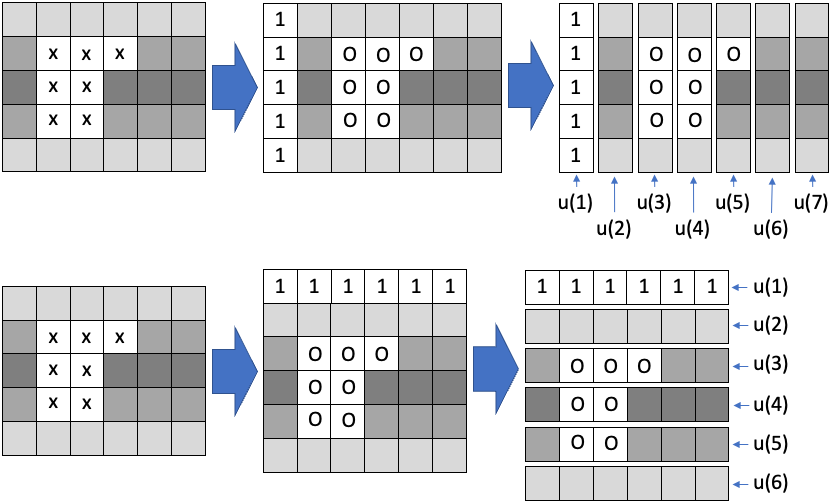}
\caption{Preprocessing of an exemplary two-dimensional input sample for feeding samples column-wise (upper part) or row-wise (lower part) into our ESN models requires adding an additional column or row of ones, respectively, as first time step $u(1)$ and replacing missing values by zero in any case. Invalid grid points, zeros and ones are marked as X, O and 1, respectively.}
\label{fig:preprocessing_ESN}
\end{centering}
\end{figure}

Our CNN models require two-dimensional input samples. Missing values are replaced by zero. Input samples are then scanned by quadratic kernels. Kernel sizes and step sizes, also referred to as strides, need to be specified. Figure \ref{fig:preprocessing_CNN} sketches on the left-hand side an exemplary sample of dimensions $5 \times 6$ grid points. Here, we assume to work with a kernel of size $3 \times 3$ and step size is also chosen to be $3$ in both directions, x and y. This requires adding a row of zeros to allow the sample to be completely scanned by our specified kernel and step size. The modified input sample is shown on the right-hand side of Figure \ref{fig:preprocessing_CNN}. Adding a row of zeros doesn't affect mean relevance maps. Zero relevance is attributed to these additional grid points when we only consider positive pre-activations $z^+$ in the propagation rule, as stated in Equation \ref{equation:LRP_propagation}. 

\begin{figure}[]
\begin{centering}
\includegraphics[width=0.7\textwidth]{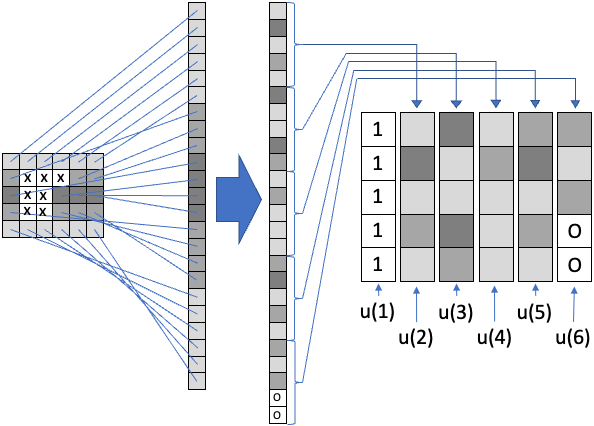}
\caption{Preprocessing of an exemplary two-dimensional input sample for feeding samples as equal-sized pieces of size 5 into our ESN models. Only valid grid points are transformed into a one-dimensional vector and randomly permuted. Need to add two zeros at the trailing edge before splitting. First time step's inputs consist of ones. Invalid grid points, zeros and ones are marked as X, O and 1, respectively.}
\label{fig:preprocessing_ESN_piecewise}
\end{centering}
\end{figure}

For our ESN models, we need to add a dummy column of ones as first input step to absorb residual relevance in any case as described above. The way we deal with missing data varies for different techniques of feeding samples into ESN models. For feeding samples column- or row-wise, missing values are set to zero, as for our CNN models. Samples are then sliced column- or row-wise, as sketched in the upper and lower part of Figure \ref{fig:preprocessing_ESN}, respectively. When feeding samples as equal-sized pieces, we only consider valid grid points. These grid points are then transformed into a one-dimensional vector, randomly permuted and split. The size or the number of grid points per piece needs to be specified. The number of valid grid points needs to be dividable by the piece size. Therefore, we optionally add zeros on the trailing edge to allow splitting into equal-sized pieces. Again, this doesn't affect mean relevance maps, since zero relevance is attributed to these grid points. For showing resulting mean relevance maps, additional grid points are discarded. Preprocessing is demonstrated in Figure \ref{fig:preprocessing_ESN_piecewise} for an exemplary sample consisting of 5 rows and 6 columns. Desired piece size is 5 in this example.

\end{document}